\pdfoutput=1

\documentclass[11pt]{article}

\usepackage[preprint]{acl}

\usepackage{times}
\usepackage{latexsym}
\usepackage{kotex}
\usepackage{tabularx} 
\usepackage{adjustbox}
\usepackage{subcaption}
\usepackage{graphicx}
\usepackage{amsmath}
\usepackage{multirow}
\usepackage{booktabs}
\usepackage{amssymb}
\usepackage{soul}
\definecolor{customblue}{rgb}{0.85, 0.847, 0.941}
\sethlcolor{customblue}
\usepackage[T1]{fontenc}

\usepackage[utf8]{inputenc}

\usepackage{microtype}

\usepackage{inconsolata}

\usepackage{graphicx}

\title{ZEBRA: Leveraging Model-Behavioral Knowledge \\for Zero-Annotation Preference Dataset Construction}

\author{
  Jeesu Jung$^{1}$, Chanjun Park$^{2\star}$, Sangkeun Jung$^{1}$\thanks{Equal contribution. Corresponding author} \\
  $^{1}$Chungnam National University, $^{2}$Korea University \\
  \texttt{jisu.jung5@gmail.com}, \texttt{bcj1210@korea.ac.kr}, \texttt{hugmanskj@gmail.com}
}

\begin{document}
\maketitle
\begin{abstract}
Recent efforts in LLM alignment have focused on constructing large-scale preference datasets via human or Artificial Intelligence(AI) annotators. However, such approaches rely on \textit{instance-wise} supervision, incurring substantial annotation cost and limited interpretability. In this paper, we propose \textbf{ZEBRA}—a \textit{model behavior-wise zero-annotation} framework that constructs preference data by leveraging model behavior knowledge derived from benchmark performances.

ZEBRA binarizes response pairs by evaluating the quality and similarity of their origin models, entirely bypassing instance-level annotation. This allows scalable, controllable, and cost-effective alignment data generation. Empirical results show that ZEBRA achieves alignment performance comparable to instance-supervised methods, despite requiring no manual or model-based labeling. 


\end{abstract}

\section{Introduction}

Aligning large language models (LLMs) with human preferences is an essential step toward making them both useful and safe. A common way to achieve this is through instance-wise labeling, where pairs of model responses are compared one by one to see which is better. Well-known methods like Reinforcement Learning from Human Feedback (RLHF\citet{rlhf}) and Artificial Inteligence(AI)-based labeling(RLAIF\citet{RLAIF}) often use this strategy.
 
However, \textit{instance-wise} labeling faces two major challenges. First, it is very costly, whether it involves human annotators or additional computational resources for LLM-based labeling\cite{diverging_preference, llm_as_a_judge}. Second, it lacks a global view of the model’s behavior\cite{hallucination}. Since each response pair is judged in isolation, it is difficult to consider broader factors. For example, whether fluency should outweigh factual accuracy. Or whether a model’s outputs are consistently aligned with certain policies\cite{helpsteer2}. This can lead to labeling noise, mistakes, and limited interpretability.

 \begin{figure}[t]
    \centering
    \includegraphics[width=\linewidth]{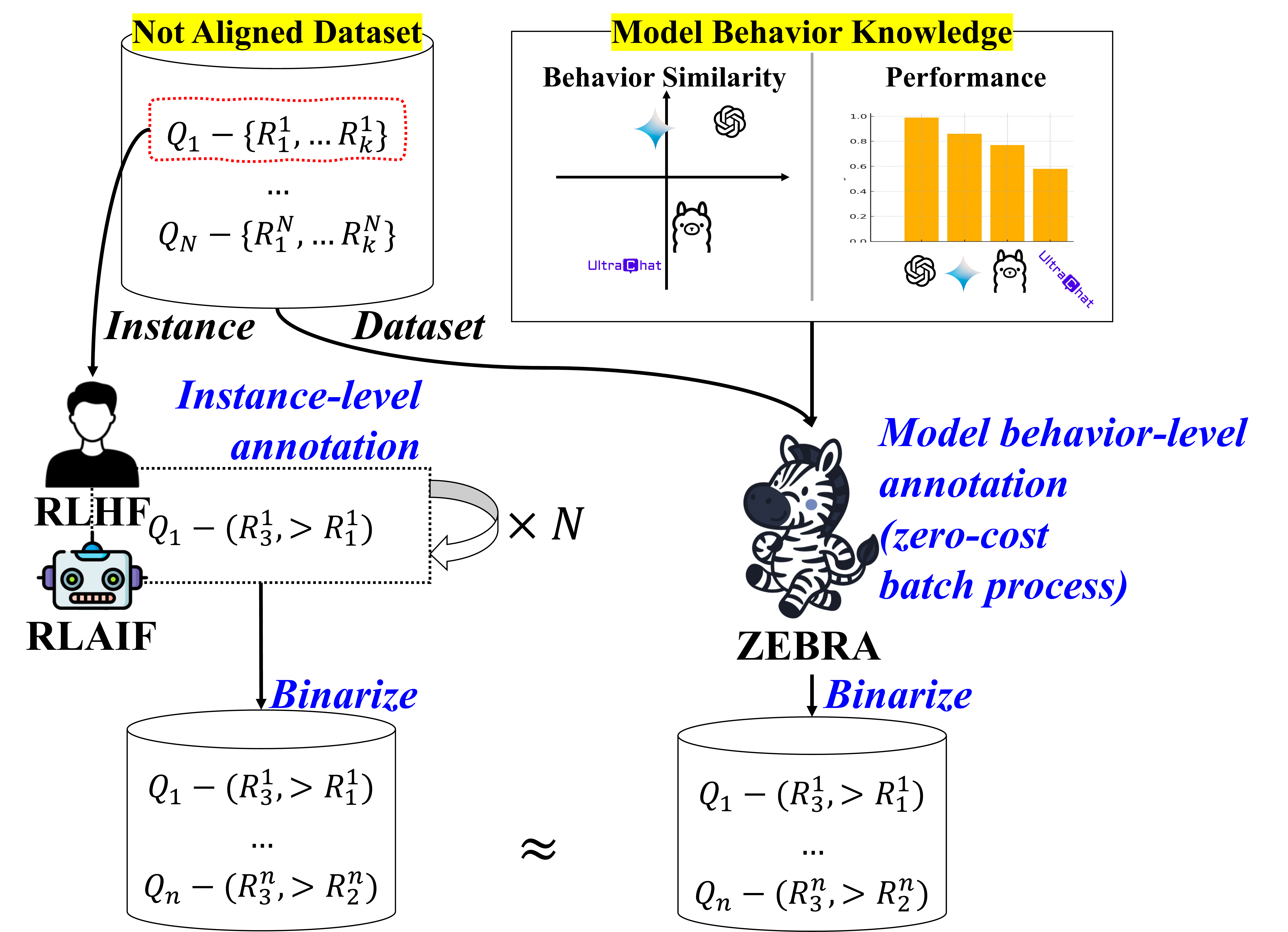}
    \caption{Human annotation\cite{rlhf} and AI labeler-based RLAIF\cite{ultrafeedback} use \textit{instance-wise} annotations, while ZEro-annotation Behavior-based Response Alignment(ZEBRA) applies \textit{model behavior-wise} annotation based on \textbf{model behavioral knowledge}, which captures proficiency and similarity across models.
    $M$ is the model, $R_i$ is the generated response from $M_i$ about Instruction $Q$. $N$ is the number of preference dataset and $K$ is the number of models}
    \label{fig:overall_flow}
    \vspace{-1.5em}
\end{figure}

To address these limitations, we propose a new preference binarization approach called \textbf{ZEro-annotation Behavior-based Response Alignment (ZEBRA)} (Figure~\ref{fig:overall_flow}). The main idea is: (1) extract each model’s behavioral patterns from its past performance trajectories, (2) measure and compare these behaviors in terms of model strength or similarity, and (3) assign preferences at the \textit{model} level rather than for each individual response pair.
 
We implement this idea through three key components. First, we define \textbf{Model Behavior Knowledge (MBK)} for LLMs (discuss in Section~\ref{sec:dataset}). Second, we propose a way to quantify and collect MBK from objective data sources such as benchmark performance. Third, we introduce three strategies—based on \textit{superiority}, \textit{similarity}, and a \textit{hybrid}—to construct a binarization dataset in a zero-annotation, cost-free manner.
 
A major advantage of our approach is that it creates response pairs based on model superiority without any additional human or LLM labeling. By classifying models with higher benchmark scores as “positive” and those with lower scores as “negative,” we can systematically label preferences throughout the dataset. This significantly reduces the cost of annotation and, because MBK visualizes each model’s behavior pattern, increases the interpretability of the preference decisions.
 
Through extensive experiments, we show that \textbf{ZEBRA} achieves performance comparable to existing instance-wise labeling methods in the Ultrafeedback\cite{ultrafeedback} dataset—\textit{without} any extra labeling cost.
 
In summary, our contributions are as follows:
\begin{itemize}
    \item We introduce ZEBRA, \textit{a zero-annotation alignment framework} that determines preferences from quantified model-level behavior, bypassing instance-level supervision.
    \item We demonstrate that the \textit{Model Behavior Knowledge (MBK)} from benchmark performance offers alignment signals comparable to instance-wise labeling.
    \item We empirically show that ZEBRA matches the performance of established methods such as RLHF and RLAIF, yet requires \textit{no additional labeling cost}.
\end{itemize}

\section{Preliminaries}

\subsection{Instance-Level Preference Construction}

Most existing preference learning frameworks for LLM alignment—such as Reinforcement Learning from Human Feedback(RLHF, \citet{rlhf}) and AI-generated preference methods (RLAIF, \citet{RLAIF})—rely on \textbf{instance-level pairwise supervision}. Given an instruction $x$, multiple candidate responses $\{r_1, r_2, ..., r_k\}$ are scored or ranked by either human annotators or automated scoring models. This generates preference tuples $(x, r_i \succ r_j)$, where $r_i$ is preferred over $r_j$ under some evaluation criteria (e.g., helpfulness, truthfulness, coherence).

The preference construction process typically involves:

\vspace{-1em}
\begin{itemize}
    \item Generating multiple responses per instruction using different models or decoding strategies.
    \item Computing preference labels via either human judgment or model-based scoring (e.g., GPT-4\cite{gpt4}).
    \item Aggregating these labels into a pairwise dataset for alignment tuning.
\end{itemize}

\vspace{-1em}
\subsection{Challenges of Instance-Level Supervision}
While instance-level supervision has proven effective in aligning LLMs, it remains costly, noisy, and difficult to scale. Despite its popularity, instance-level supervision suffers from three core limitations:

\vspace{-1em}
\begin{enumerate}
    \item \textbf{Costly evaluation}: Annotating or scoring each response pair requires substantial human or computational effort.
    \item \textbf{Preference triviality}: When candidate responses differ significantly in quality, the preference label becomes trivial, contributing little to alignment learning.
    \item \textbf{Instruction-level variance}: Difficulty and ambiguity in the instruction $x$ can introduce noise into the preference signal, especially for automated labelers.
\end{enumerate}

These limitations underscore the need for alternative approaches that construct preference signals without relying on per-instance scoring.

\vspace{0.5em}
\subsection{Motivation for Model Behavior-Level Preference}

We propose that instead of relying on per-instance evaluation, one can leverage the \textbf{intrinsic capabilities of the response-generating models} themselves. As models exhibit differences in core competencies measurable via standardized benchmarks, we hypothesize this can replace instance-level annotations. 

\begin{figure*}[t]
    \centering
    \includegraphics[width=0.9\linewidth]{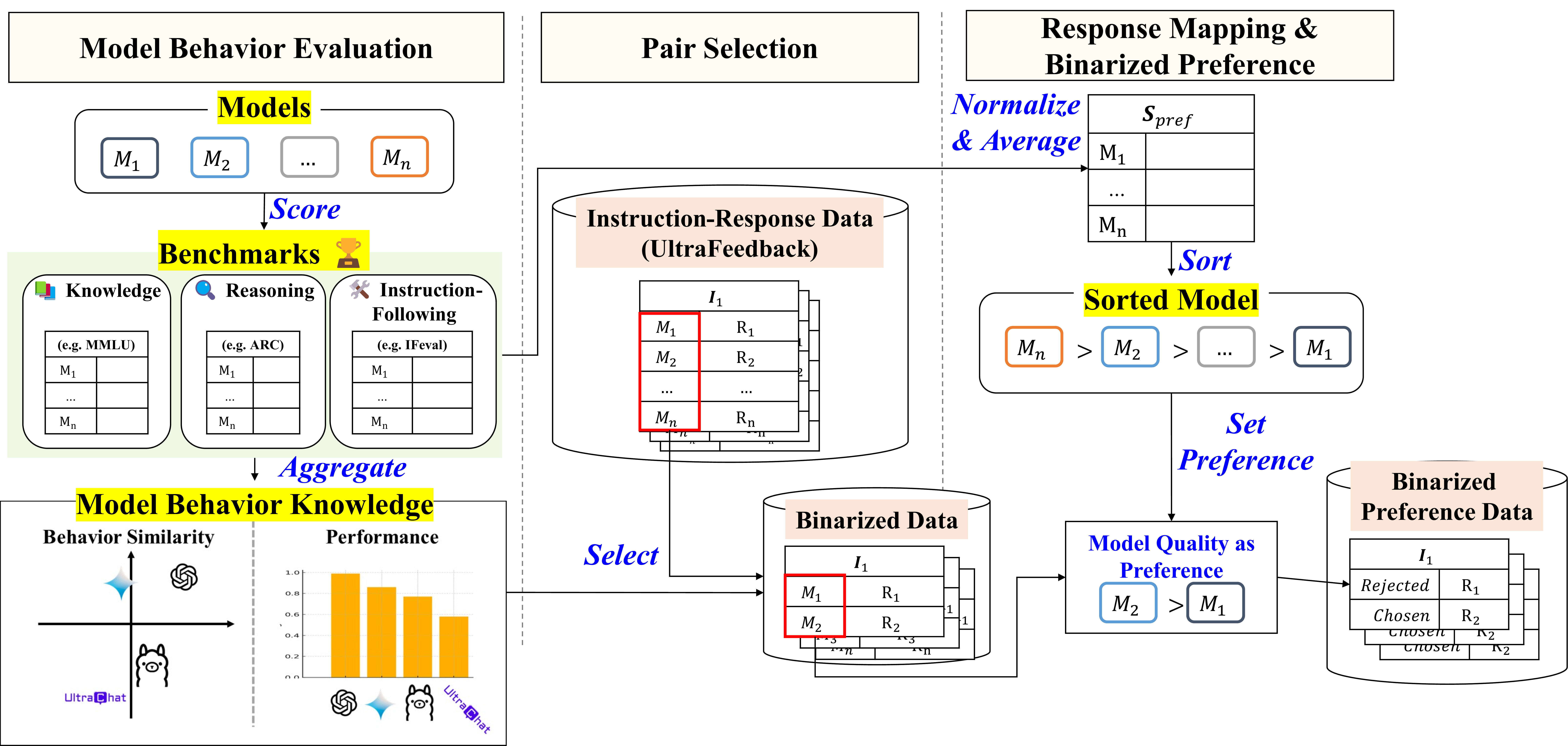}
    \caption{Editing process of ZEBRA framework. $n$ is the number of models. $\{M_1, M_2, ..., M_n\}$ represents the total set of candidate models. $\{I_1, I_2, ..., I_n\}$ represents the instruction input of dataset. $\{R_1, R_2, ..., R_n\}$ represents the total responses from the candidate models. $R_x$ and $R_y$ are the chosen and rejected responses. $S_{pref}$ score summarizes the model's benchmarks ability}
    \label{fig:overall_structure}
    \vspace{-1em}
\end{figure*}

\section{ZEro-annotation Behavior-based Response Alignment Framework}

\subsection{Instance-level Annotation vs. \\Model behavior-level Annotation}
Traditional instance-level binarization relies heavily on detailed, provided by human or AI annotators\cite{diverging_preference, RLAIF_eval}. Each pairwise comparison demands significant effort and resources to maintain consistency and interpretability. Such an approach often results in annotation noise, limited scalability, and considerable expense.

In contrast, our proposed ZEBRA framework leverages intrinsic behavioral knowledge derived from model performance across various benchmarks, ZEBRA systematically matches and pairs responses. Responses from models with proven higher competencies form positive labels, while those from models with lower competencies become negative labels. This innovative model-level approach drastically reduces annotation costs, minimizes noise, and enhances scalability. Additionally, the behavior similarity-based matching provides explicit control over the difficulty and nuance of preference comparisons, leading to clearer and more meaningful alignment outcomes.

\subsection{Model Behavior Knowledge}
While most preference-learning pipelines focus on differences between individual responses, our approach highlights Model Behavior Knowledge (MBK)—the comprehensive record of each model’s past behaviors and capabilities. We define MBK using two sets of metrics:
 \begin{itemize}
     \item  \textbf{Superiority: } How much better (or worse) a model is compared to others, based on overall or task-specific proficiency.
     \item  \textbf{Similarity: } How likely a model is to behave similarly to other models.
 \end{itemize}
 
These metrics provide a principled basis for quantifying both a model’s general strength and its behavioral proximity to its peers. Within a preference-learning pipeline, \textit{superiority} functions as a global preference signal: when one model consistently outperforms another across standardized benchmarks, its responses are considered preferable overall.

Conversely, \textit{behavioral similarity} facilitates the systematic construction of challenging comparison sets. When two models are behaviorally similar—for example, they exhibit comparable reasoning performance—their responses become difficult to distinguish. Training on such hard-to-distinguish pairs guides the preference learner to focus on subtle qualitative differences, resulting in more nuanced and robust alignment.

\subsection{Model Behavior Evaluation using Benchmark Performances}
To capture MBK in a practical, objective way, we rely on external benchmark performance data for each LLM. Many models are already evaluated across diverse, standardized tasks (e.g., reasoning, factual accuracy, instruction-following). We aggregate these benchmark scores to form each model’s MBK profile.

Benchmark performance offers several advantages for extracting MBK:
\begin{itemize}
    \item It provides \textit{reliable metrics} for the core competencies of large language models.
    \item It enables \textit{straightforward comparison and aggregation} across multiple models.
    \item Because benchmark scores are published and fixed at release, they impose no constraints on subsequent \textit{data or model expansion}.
\end{itemize}

\begin{figure}[t]
    \centering
    \includegraphics[width=\linewidth]{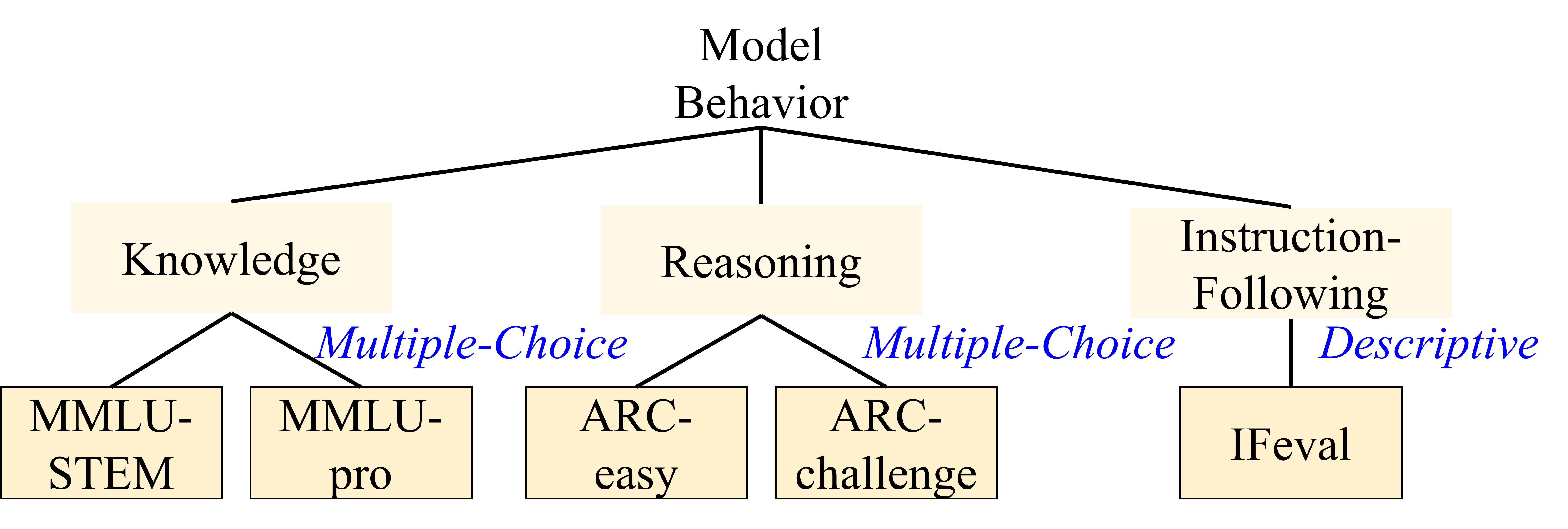}
    \caption{Selected benchmarks for evaluating the models behavior knowledge: knowledge, reasoning, and instruction-following capabilities}
    \label{fig:benchmark}
    \vspace{-1.7em}
\end{figure}

The list of benchmarks used in our analysis is provided in Figure~\ref{fig:benchmark}. 

\begin{figure}[t]
    \centering
    \includegraphics[width=\linewidth]{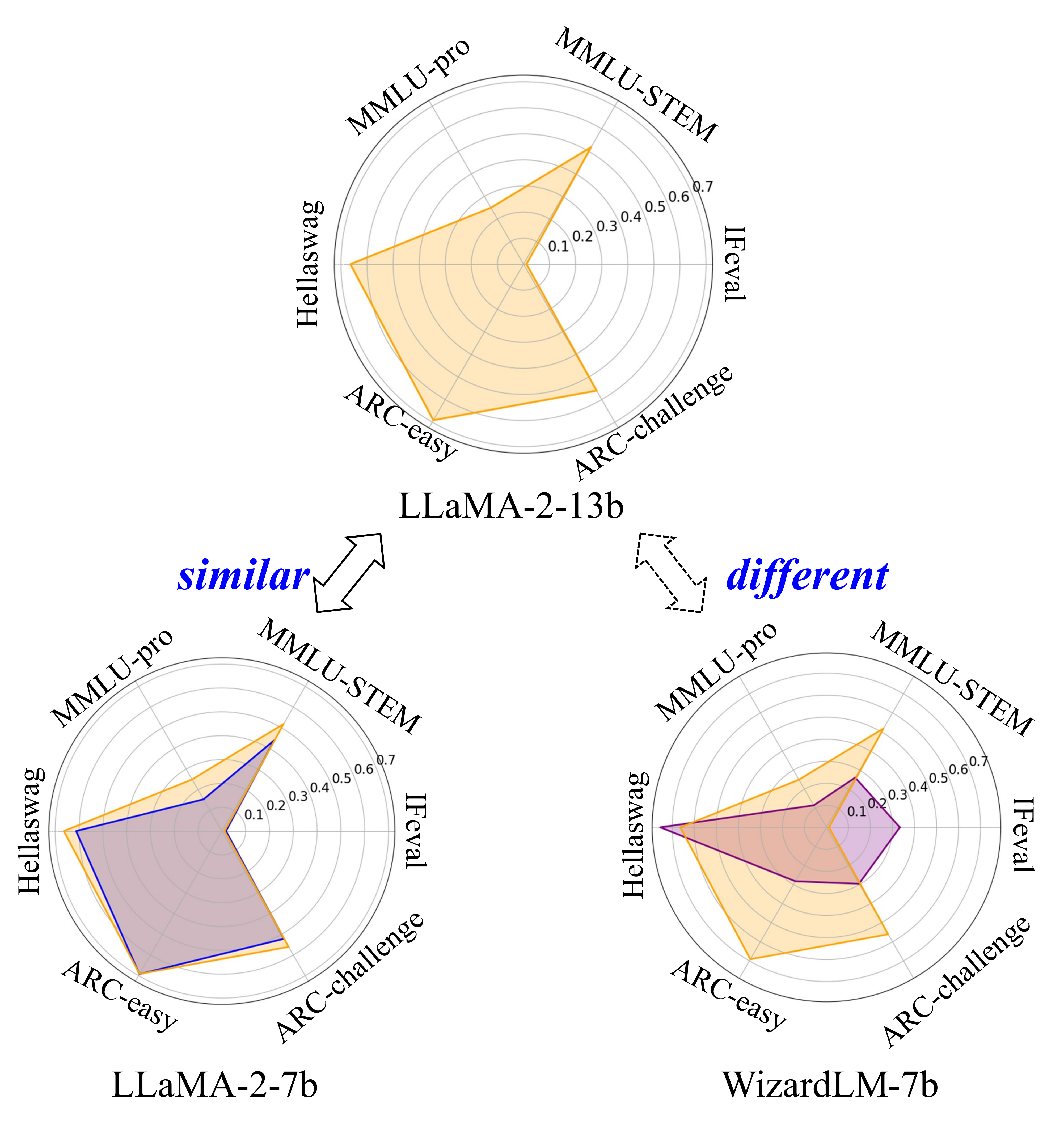}
    \caption{Hexagonal radar plots visualizing model ability profiles. Models judged as \textit{similar} exhibit comparable benchmark performance trajectories, while \textit{different} models show clearly divergent patterns.}
    \label{fig:example_similar_different_model}
    \vspace{-1.5em}
\end{figure}

Figure~\ref{fig:example_similar_different_model} illustrates how the behavioral similarities inferred from benchmark performance can reflect actual model similarities. For example, the LLaMA-2-13b model exhibits a pattern closely resembling that of its 7b counterpart. In contrast, WizardLM-7b demonstrates a markedly different behavioral trajectory.

We define a model's ability vector $v_i \in \mathbb{R}^m$ across $m$ benchmark tasks:
\[
v_i = \left[ s_{i}^{(1)}, s_{i}^{(2)}, \dots, s_{i}^{(m)} \right],
\]
where $s_{i}^{(b)}$ is the normalized score of model $M_i$ on benchmark $b$, reflecting its relative capability in a specific behavioral dimension (e.g., knowledge, reasoning, instruction-following). These standardized behavior vectors serve as the foundation for both quality-based and similarity-based anchoring, enabling \textbf{zero-annotation binarization} of preference pairs without per-instance supervision.


\subsection{Benchmark-based Model Behavior Quantification}
To effectively binarize preference data, it is essential to quantify model behavior from two complementary perspectives: \textit{behavior quality} and \textit{behavior similarity}. ZEBRA leverages benchmark-derived measures of these aspects to systematically pair positive responses with suitable negative counterparts. By quantifying both the absolute competency of individual models and the relative similarity between models, ZEBRA ensures that each positive-negative pair captures meaningful contrasts in model capabilities, thus maximizing alignment informativeness.

\textbf{Model Behavior Superiority (MB-SUP)} quantifies the overall behavioral competency of model $M_i$ as the aggregate of its normalized benchmark scores:
\begin{equation}
\text{MB-SUP}(M_i) = \frac{1}{m} \sum_{k=1}^{m} s_i^{(b)}.
\end{equation}
This scalar value serves as the ranking basis for constructing \textit{Behavioral Superiority Anchors}.

\textbf{Model Behavior Similarity (MB-SIM)} between models $M_i$ and $M_j$ is defined as the similarity between their behavior vectors:
\[
\text{MB-SIM}(M_i, M_j) = \text{similarity}(v_i, v_j).
\]
Higher values indicate stronger alignment in general-purpose capabilities, and MB-SIM serves as the criterion for selecting comparable model pairs in \textit{Behavioral Similarity Anchoring}.

\begin{figure*}
    \centering
    \includegraphics[width=\linewidth]{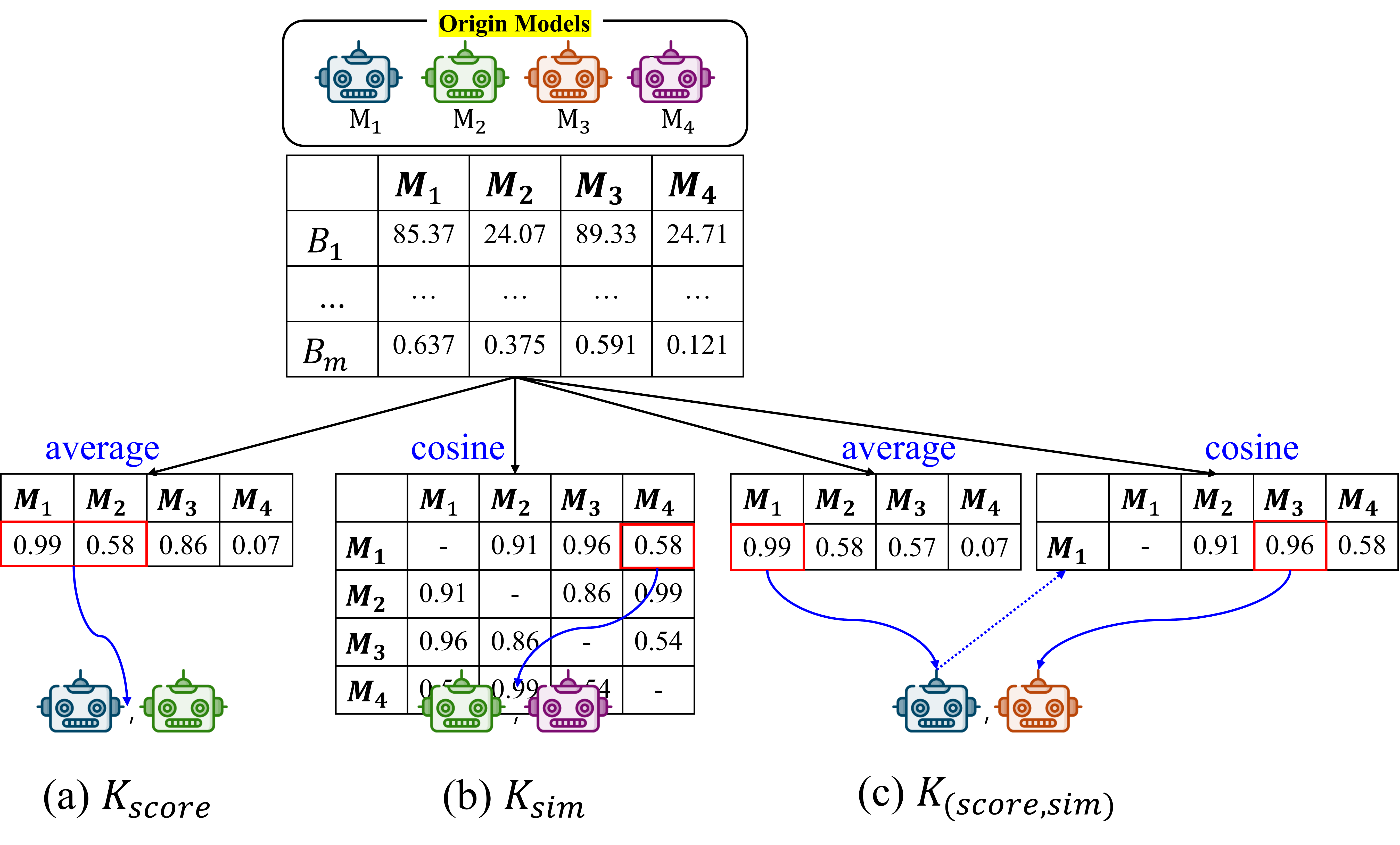}
    \caption{Anchoring strategies symbolic aliases and the example of strategy SUP, SIM and SUP+SIM with model and labeled preference.}
    \label{fig:example_kinship}
    \vspace{-1.5em}
\end{figure*}

\subsection{Strategy of Preference Binarization}

ZEBRA introduces multiple strategies to systematically convert benchmark-derived model behaviors into binary preference pairs. Based on how MB-SUP and MB-SIM are utilized, our strategies can be clearly categorized as follows on Figure~\ref{fig:example_kinship}.

We detail each strategy below:

\vspace{0.5em}
\noindent\textbf{Strategy 1: Superiority-first Anchoring (SUP)}\\
In this strategy, we explicitly select the top-two models based on their MB-SUP scores: the highest-scoring model (top-1) and the second-highest-scoring model (top-2). Responses from the top-1 model serve as positive anchors, while responses from the top-2 model become negative counterparts. This approach emphasizes explicit quality distinctions, clearly defining superior responses and ensuring meaningful, informative alignment contrasts.

\vspace{0.5em}
\noindent\textbf{Strategy 2: Similarity-first Anchoring (SIM)}\\
Responses from models sharing similar behavioral patterns (high MB-SIM) are paired first. Within these pairs, the response from the model with higher MB-SUP is selected as the anchor (positive response). This strategy emphasizes behavioral similarity, enhancing nuanced alignment comparisons.

\vspace{0.5em}
\noindent\textbf{Strategy 3: Hybrid Anchoring (SUP+SIM)}\\
Model pairs are selected by simultaneously considering both MB-SUP and MB-SIM criteria. This balanced approach ensures each response pair reflects meaningful contrasts in model quality, while maintaining behavioral similarity for refined granularity.

These strategies enable ZEBRA to flexibly and effectively tailor preference construction according to the desired granularity, alignment objectives, and computational resources. Figure~\ref{fig:example_kinship} represent the example of each strategy.

\subsection{Zero-Annotation Preference Construction}
Given a set of instructions $\mathcal{X}$ and a pool of response-generating models $\mathcal{M}$, the construction proceeds as follows:

\begin{enumerate}
    \item \textbf{Model Behavior Evaluation:} Each model $M_i \in \mathcal{M}$ is benchmarked across $m$ tasks to obtain model’s ability vector $v_i$.
    \item \textbf{Pair Selection:} Using a chosen strategy, a set of model pairs $\mathcal{P} = \{(M_i, M_j)\}$ is selected where $\text{MB-SIM}(M_i, M_j) \geq \tau$. we set $\tau$ at 0.1, as cosine similarity values below this threshold indicate a lack of meaningful similarity between models.
    \item \textbf{Response Mapping:} For each instruction $x \in \mathcal{X}$, the corresponding responses $\{r_i, r_j\}$ from $(M_i, M_j)$ are retrieved.
    \item \textbf{Preference Assignment:} A binary label is assigned via:
    \[
    \text{Pref}(r_i, r_j) =
    \begin{cases}
    1 & \text{if } S_{\text{pref}}(M_i) > S_{\text{pref}}(M_j), \\
    0 & \text{otherwise}.
    \end{cases}
    \]
\end{enumerate}


\vspace{-1em}
This pipeline constructs a binarized preference dataset at scale without any manual or per-instance scoring. Figure~\ref{fig:overall_structure} shows the total process of the pipeline.

\section{Experimental Setup}
In this paper, we propose the ZEBRA framework. To validate the functionality of this framework and the characteristics of Alignment Tuning, we conducted experiments to address the following Research Questions (RQs):  
\begin{itemize}  
   \item \textbf{RQ1}: Does MB-SIM represent the models' similarity?
   \item \textbf{RQ2}: Does strategies affect the binarization process?  
   \item \textbf{RQ3}: Does ZEBRA reduce annotation cost and computational overhead compared to instance-level methods?
\end{itemize}  

\subsection{Alignment Dataset}\label{sec:dataset}
For preference binarization, we utilized the UltraFeedback dataset \cite{ultrafeedback}, which includes diverse responses generated by both commercial and open-source language models. It provides multiple model responses, making it suitable for constructing a strong RLAIF baseline, for which we adopted the original score aggregation method. In contrast, ZEBRA uses only the response pool without referencing any scores, ensuring a clean comparison focused on the binarization strategy. 

\textbf{Model coverage.} UltraFeedback contains outputs from 17 LLMs: GPT-4\cite{gpt4}, GPT-3.5 Turbo\cite{InstructGPT}, and Bard\cite{bard}. Additionally, several models from the Llama family were included, such as Llama-2 (7B, 13B, and 70B)-chat\cite{llama2}, UltraLM-13B\cite{ultralm}, WizardLM (7B, 13B, and 70B)\cite{wizardlm}, Vicuna-33B\cite{vicuna}, and Alpaca-7B\cite{alpaca}. Beyond the Llama-based architectures, the dataset also features responses from other notable models, including Falcon-40B-instruct\cite{falcon40b}, MPT-30B-chat\cite{MPT}, StarChat-Beta\cite{starchat}, and Pythia-12B\cite{pythia}. 

Although the Ultrafeedback dataset contains results from UltraLM-65B\cite{ultralm}, its performance could not be accurately assessed. To maintain the reliability of our evaluation, we excluded these results from the dataset composition.

\vspace{-0.5em}
\subsection{Models}\label{sec:model}
We fine-tuned and evaluated Llama-3.1-(3B, 8B)-Instruct\cite{llama3} and Qwen2.5-(3B, 8B)-Instruct\cite{qwen25} on the ZEBRA-binarized dataset.  
This setup enables cross-family comparison and quantifies the alignment gains from behavior-aware preference construction.

\subsection{Training Algorithm for Alignment Tuning}
We tested two different learning method for alignment tuning using the ZEBRA Framework:
\vspace{-0.7em}
\begin{itemize}
    \item Supervised Fine-Tuning (SFT)
    \item Direct Preference Optimization (DPO) \cite{dpo}
\end{itemize}
\vspace{-0.7em}
The implementation details are provided in Appendix \ref{app:implementation_detail}.
The full source code, benchmark results, and preference binarization scripts will be released after publication via our project repository under the MIT License. The snapshot corresponding to this paper is version-tagged.

\section{Experimental Results}

\begin{figure}[t]
    \includegraphics[width=1\linewidth]{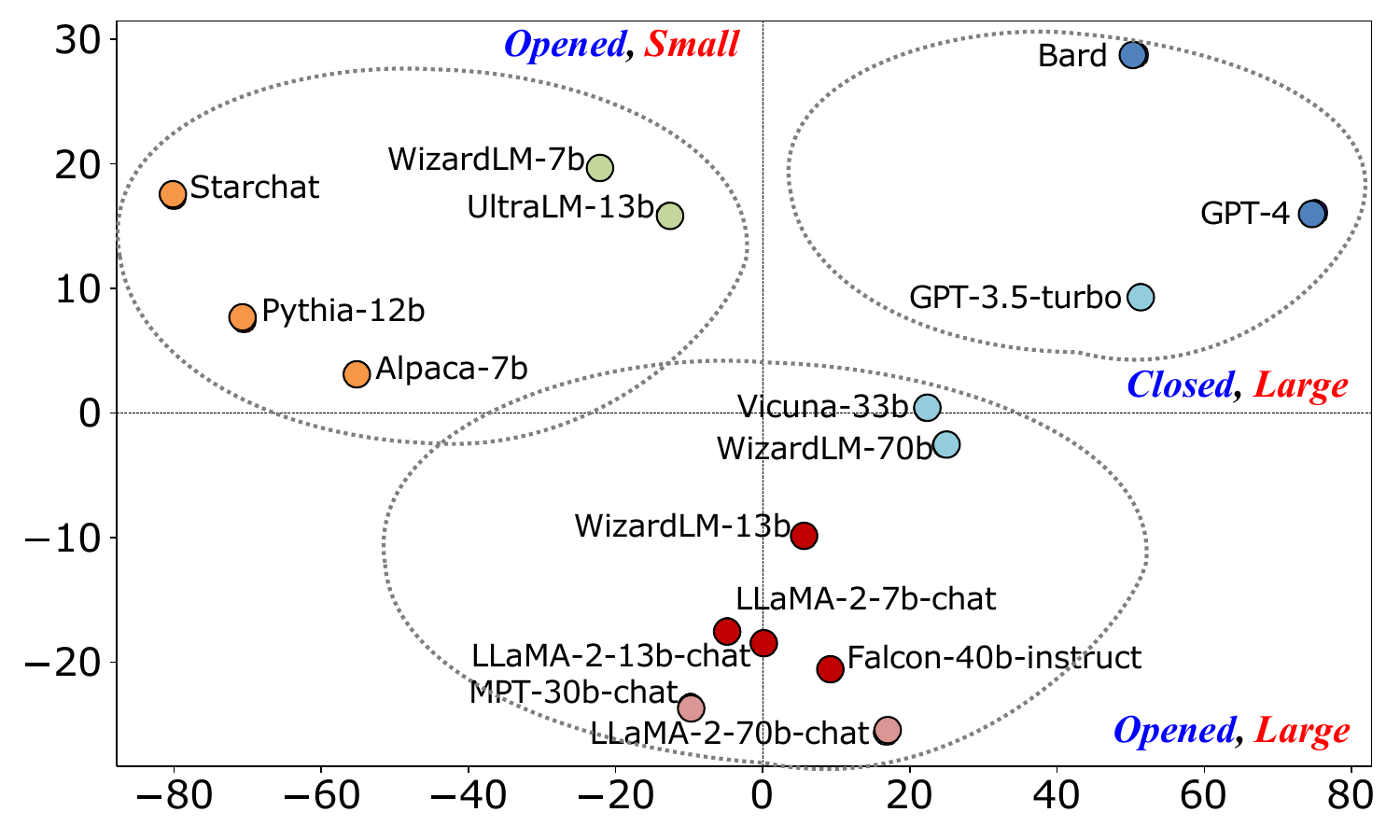}
    \caption{PCA-based visualization of relationships among models\cite{PCA}. The axes reflect similarity-based variance.
    Similar models are positioned closer to each other. Circle colors indicate clusters identified by hierarchical clustering.}
    \label{fig:Q1_kinship_vis}
    \vspace{-1em}
\end{figure}

\begin{table*}
    \centering
\begin{adjustbox}{width=\linewidth}
    \begin{tabular}{c|c|c|cc|cc|c}
        \toprule
        & & & \multicolumn{2}{c|}{Knowledge} & \multicolumn{2}{c|}{Reasoning} & Instruction-Following \\
        \cline{3-7}
        Category & Strategy & Average & MMLU & MMLU-Pro & ARC-Easy & ARC-Challenge & IFeval \\
        \hline
        Baseline & RLAIF & \textbf{0.31} & \textbf{0.36} & \textbf{0.15} & 0.40 & \textbf{0.40} & 0.28 \\
        \hline
        \multirow{3}{*}{ZEBRA(Ours)}  
            & SUP & \hl{\textbf{0.31}}(-0.00) & 0.33 & \hl{\textbf{0.15}} & \hl{0.40} & 0.37 & \hl{\textbf{0.30}} \\
            & SIM & 0.29(-0.02) & 0.34 & \hl{\textbf{0.15}} & 0.36 & 0.33 & \hl{0.29} \\
            & SUP+SIM & 0.29(-0.02) & 0.30 & 0.14 & \hl{\textbf{0.41}} & 0.39 & 0.23 \\
        \bottomrule
    \end{tabular}
\end{adjustbox}
    \caption{Performance Comparison Between Instance-wise Binarized Data (Baseline) and Model Behavior-wise Binarized Data (Ours). The baseline corresponds to instance-wise scored RLAIF \cite{ultrafeedback}. The \hl{highlighted} cells indicate performance that is equal to or higher than the baseline. \textbf{Bold} text shows the best performance on the benchmarks.}
    \label{tab:performance_comparison}
\vspace{-1em}
\end{table*}

\begin{table}[t]
\centering
\begin{adjustbox}{width=\linewidth}
\begin{tabular}{lcccc}
\hline
Comparison & $t$ & $p_\text{t}$ & Mean Δ & $p_\text{perm}$ \\
\hline
SUP vs BL  & -0.3541     & 0.7282     & -0.0066    & 0.7345 \\
SIM vs BL   & -0.4374     & 0.6680     & -0.0047    & 0.7585 \\
SUP+SIM vs BL & -1.4472    & 0.1684     & -0.0296    & 0.2011 \\
\hline
\end{tabular}
\end{adjustbox}
\caption{\small Paired $t$-test and permutation test (10,000 shuffles) comparing each anchoring strategy with the baseline (BL).}
\label{tab:stat_comparison}
\vspace{-1.5em}
\end{table}

\begin{table*}[t]
\centering
\begin{adjustbox}{width=\textwidth}
\begin{tabular}{clccc}
\toprule
Category & Method & Pairs / Units & Unit Cost (USD) & Total Cost (USD) \\
\midrule
\multirow{3}{*}{\shortstack{Instance-wise\\RLAIF}} 
  & UltraFeedback\cite{ultrafeedback} & 64,000 & 0.252 & 16,128\\
  & Safer-Instruct\cite{safer_instruct} & 10,254 & 0.063 & 646\\
   & OpenHermesPreferences\cite{open_hermes_preferences} & 989,000 & 0.126 & 124,614\\
\midrule
Model behavior-wise & ZEBRA (ours) & 64,000 & 0 & 0 \\
\bottomrule
\end{tabular}
\end{adjustbox}
\caption{Labeling‐cost comparison between LLM‐labeled RLAIF datasets, and our benchmark-table approach. 
cost is estimated assuming equivalent labor per rating.}
\label{tab:labeling-cost}
\vspace{-1.5em}
\end{table*}
\subsection{RQ1: Dataset Reconstruction using ZEBRA}
The proposed ZEBRA framework enables the application of a unified Total Ranking map across the entire dataset, facilitating a structured and consistent preference mapping process. Utilizing this reconstructed preference dataset, we conducted model training while ensuring that each critic’s binarized response was systematically incorporated. The effectiveness of this reconstructed dataset was assessed by evaluating the trained models on standardized benchmarks. The detailed results of these evaluations are presented in Appendix \ref{app:benchmarks}.

For evaluation, we employed prediction-based assessment methodologies across all benchmark tasks. Specifically, for ARC, MMLU, and MMLU-Pro, we adapted the MMLU-Pro evaluation framework, modifying only the multiple-choice options to align with our dataset. For IFeval, we leveraged its native evaluation framework to ensure consistency in assessment.

To analyze the relationships between models, we computed SUP and SIM by normalizing evaluation results across six benchmark tasks. Figure \ref{fig:Q1_kinship_vis} presents a Principal Component Analysis (PCA)\cite{PCA} visualization of these relationships, illustrating distinct clustering patterns among models. Generally, smaller-scale models tend to cluster in the first quadrant, models with closer Model Behavior relationships in the second quadrant, while Llama-based models and models exceeding 10B parameters are predominantly distributed in the third and fourth quadrants. This distribution underscores the critical role of model training data, training algorithms, and scale in determining Model Behavior relationships among models.

We further verify that anchor pairs selected by SIM indeed yield semantically closer responses. 
Using a TF–IDF\cite{tf_idf}, we obtain an average response-pair similarity of 0.4623 for the \emph{most similar model pair} ($\operatorname{MB-SIM}\uparrow$) versus 0.4129 for the \emph{least similar model pair} ($\operatorname{MB-SIM}\downarrow$). This 12\% relative gap confirms that high-MB-SIM anchoring indeed surfaces finer-grained yet coherent preference signals.

These findings highlight the effectiveness of the ZEBRA framework in reconstructing preference datasets, providing a more structured and informative approach to model alignment.


\subsection{RQ2: Performance of ZEBRA Binarization}
To assess the effectiveness of ZEBRA binarization, we examined whether model performance can serve as an indicator of data quality. 
The results of this comparison are presented in table \ref{tab:performance_comparison}. Across all benchmark tasks, the Model Behavior-based scoring metric, SUP, demonstrates performance levels nearly equivalent to the instance-wise RLAIF method, with a minimal deviation of only 0.008. 

Notably, for MMLU-Pro and IFeval, ZEBRA-based binarization even surpasses the RLAIF baseline by approximately 0.01, indicating that structured preference mapping via Model Behavior can yield competitive or superior alignment outcomes. Table \ref{tab:performance_comparison} shows the results of the average scores for each methodology. The detailed results, including average scores for each methodology, are summarized in Appendix \ref{app:total_performance}.


\paragraph{Statistical significance.}
Table~\ref{tab:stat_comparison} reports paired $t$-tests and permutation tests (10,000 shuffles) that compare each anchoring strategy with the baseline instance-wise RLAIF data across the same 1 024 evaluation prompts. None of the strategies achieves statistical significance at $\alpha{=}0.05$,
although the hybrid SUP+SIM shows the largest absolute gain (${-}0.0296$, $p_\text{t}{=}0.17$, $p_\text{perm}{=}0.20$).
These numbers indicate that ZEBRA’s zero-annotation binarization maintains baseline-level alignment quality (within 3 pp) without incurring per-instance scoring cost, even if a decisive improvement is not yet observed.





\subsection{RQ3: Cost \& Efficiency Analysis}
\label{sec:cost}

Table~\ref{tab:labeling-cost} contrasts the \emph{labeling cost} of ZEBRA with canonical RLHF and RLAIF pipelines.  
The standing out points is below:
 \textbf{Absolute cost gap.}  
 LLM‐annotated RLAIF corpora lower this to \$0.6–\$4\,K by outsourcing each pairwise judgment to GPT‐4, yet they \emph{still purchase every label}.  ZEBRA, by reusing benchmark leaderboards, pays \textbf{no} marginal cost (\$0) for preference construction.
\textbf{Relative efficiency.}   Normalised per comparison, GPT‐4 labels ($\approx$\$0.063) are $\approx$\,\(\times10\) cheaper than human labels ($\approx$\$0.67), but ZEBRA is \emph{orders of magnitude} cheaper than both because it dispenses with pairwise annotation altogether.

Cost matters only if quality survives.  Despite a zero‐dollar label budget, ZEBRA \emph{matches or surpasses} RLAIF baseline (Table~\ref{tab:performance_comparison}); the mean performance difference across six benchmarks is $\le 0.02$.  

Consequently, ZEBRA offers a \textbf{cost‐minimal, scalable, and annotation‐free} route to high‐quality preference data. Because the price of human labor or GPT‐4 tokens scales linearly with data volume, traditional pipelines become progressively more expensive as models, tasks, and safety domains proliferate.  ZEBRA decouples alignment from annotation cost: adding a new model needs no further labels, and incorporating an extra benchmark simply augments the behavior matrix.

\section{Conclusion}

We presented \textbf{ZEBRA}, a zero-annotation framework that replaces \textit{per-instance} preference labeling with \textit{model-behavior knowledge} distilled from public benchmark tables.  By pairing responses according to either behavioral superiority or similarity—or a hybrid of the two—ZEBRA constructs 64k high-quality preference pairs at zero marginal cost, completely eliminating the dominant expense in RLHF or RLAIF pipelines.

Experiments across 6 standardized benchmarks show that models fine-tuned with \textit{SUP} achieve performance on par with, and sometimes better than, an RLAIF baseline. SIM and the hybrid SUP+SIM remain within a small, statistically insignificant margin of the baseline. These results confirm that benchmark-derived behavioral knowledge can serve as an effective proxy for preference supervision. Crucially, ZEBRA reduces marginal labeling cost to zero, reducing tens of thousands of dollars otherwise required for large-scale preference annotation.

Future work will explore richer behavioral axes such as safety, toxicity, and multilingual ability; investigate adaptive schedules that incorporate additional refined pairing strategies during training; and evaluate how well ZEBRA preferences correlate with explicit human judgments.


\section{Related Work and Background}
\subsection{Alignment Tuning}
Alignment tuning has become a central focus in enhancing LLMs to meet user expectations and ethical standards\cite{sage_rt}. 
Various preference-based learning techniques, particularly reinforcement learning methods\cite{ppo, dpo, orpo}, have been developed to facilitate this tuning process. 

These methods rely on preference datasets, which typically contain pairs of responses generated by models based on given instructions, with each pair ranked according to human or model-based evaluations. Prominent alignment datasets like \textit{Ultrafeedback\cite{ultrafeedback}}, which gathers extensive human feedback, and \textit{HH-RLHF\cite{hh_rlhf}}, which uses human-annotated preferences, provide foundational resources for alignment. To minimize the reliance on labor-intensive data curation, recent automated approaches have been introduced to filter or regenerate preference data based on specific criteria, promoting data consistency and scalability. However, such approaches often lack nuanced control over data quality, as they fail to consider the role of the origin model’s capabilities in shaping alignment effectiveness \cite{safer_instruct}.

\subsection{Limitations in Existing Preference Data Approaches}
Effective preference data construction requires a clear, rigorous set of criteria to ensure alignment quality across generated response pairs. Common criteria, including \textit{Reasoning}, \textit{Truthfulness}, and \textit{Instruction-Following}, guide the selection of data that aligns with key ethical and functional standards\cite{ultrafeedback, hh_rlhf}. High-performing models, capable of producing responses that meet these standards, are often evaluated using benchmarks like \textit{ARC\cite{arc}}, \textit{MMLU\cite{mmlu}}, and the \textit{Instruction-Following eval\cite{ifeval}} suite, which assess a model’s factual accuracy, reasoning ability, and compliance with instructions. 

The \textit{ZEBRA Framework} addresses this limitation by introducing a model behavior-level approach that emphasizes model compatibility in preference data selection. By curating preference pairs based on model behavior knowledge with similar core competencies in knowledge, reasoning, and instruction-following—the ZEBRA Framework enhances alignment coherence and ensures stable, high-quality data. 

\section*{Limitation}
In this paper, we demonstrated that ZEBRA can achieve performance comparable to existing RLAIF without requiring annotations for SFT and DPO. However, due to resource and time constraints, we were unable to validate our approach across a broader range of alignment tuning techniques. Further evaluation is needed for methods.

Additionally, our evaluation primarily focused on fundamental abilities such as Reasoning, Knowledge, and Instruction-Following. However, we did not assess ZEBRA’s performance on other important values, including factuality and ethical standard. Future work should incorporate evaluations reflecting these aspects.



\bibliography{custom}

\appendix
\section{Implementation Details for Model Training}\label{app:implementation_detail}
The model was trained for a single epoch using bfloat16 (bf16) quantization, which optimizes memory efficiency while preserving numerical precision. The training configuration incorporated the following hyperparameters: a per-device batch size of 6, gradient accumulation steps set to 4, and a learning rate of $5 \times 10^{-5}$, with 500 warm-up steps to facilitate stable convergence. 

Training was conducted on an L40 4-GPU setup, leveraging an optimized deep learning framework to enhance computational efficiency. The training pipeline focused on performance optimization through checkpointing and logging, without intermediate model evaluation during training.

\section{Prompt template for Evaluation Benchmarks}
\label{app:prompt}
\begin{figure}
    \centering
    \includegraphics[width=\linewidth]{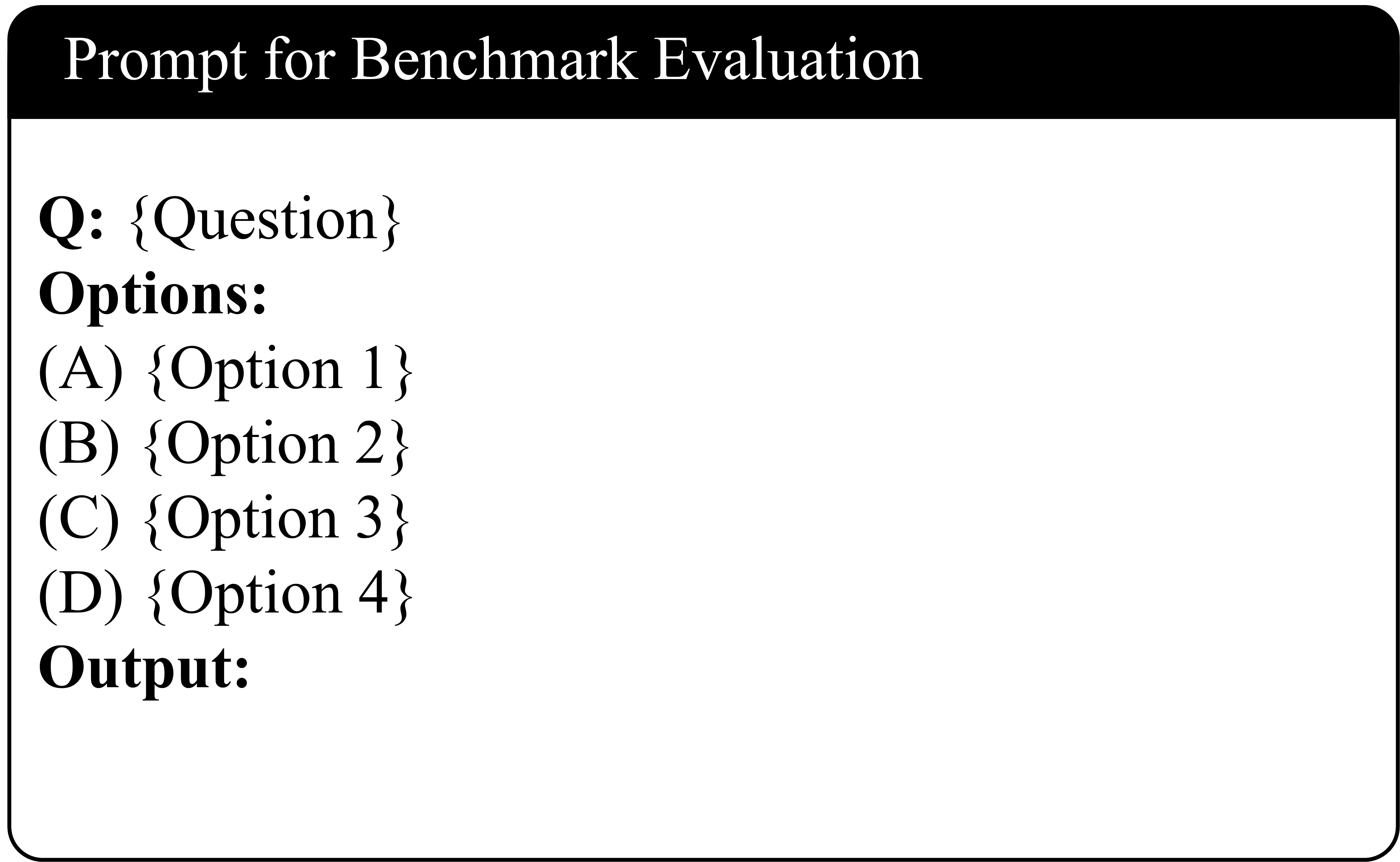}
    \caption{Evaluation template for multiple-choice benchmark evaluation. The descriptive benchmark (e.g., IFeval) was evaluated using the same template, excluding the option part.}
    \label{fig:prompt_template}
\end{figure}

To determine behavior knowledge between models, this paper evaluates benchmark performance and compares the similarity of these numerical results. The benchmark performance of all 17 models considered in this study is presented in Table \ref{fig:prompt_template}. For the start and end tokens, the template recommended in the paper was used.

\section{Evaluation Model Benchmark Performance}
\label{app:benchmarks}
To determine behavior knowledge relationships between models, this paper evaluates benchmark performance and compares the similarity of these numerical results. The benchmark performance of all 17 models considered in this study is presented in Table \ref{tab:benchmark_performance}. The actual calculation example for \( MB-SUP \) and \( MB-SIM \) can be found in Figure \ref{fig:example_sim}.

Due to the unavailability of Bard, its performance metrics have been substituted with those of Gemini-1.5-Flash.

\begin{figure*}[t]
    \centering
\begin{subfigure}{0.45\linewidth}
    \includegraphics[width=\linewidth]{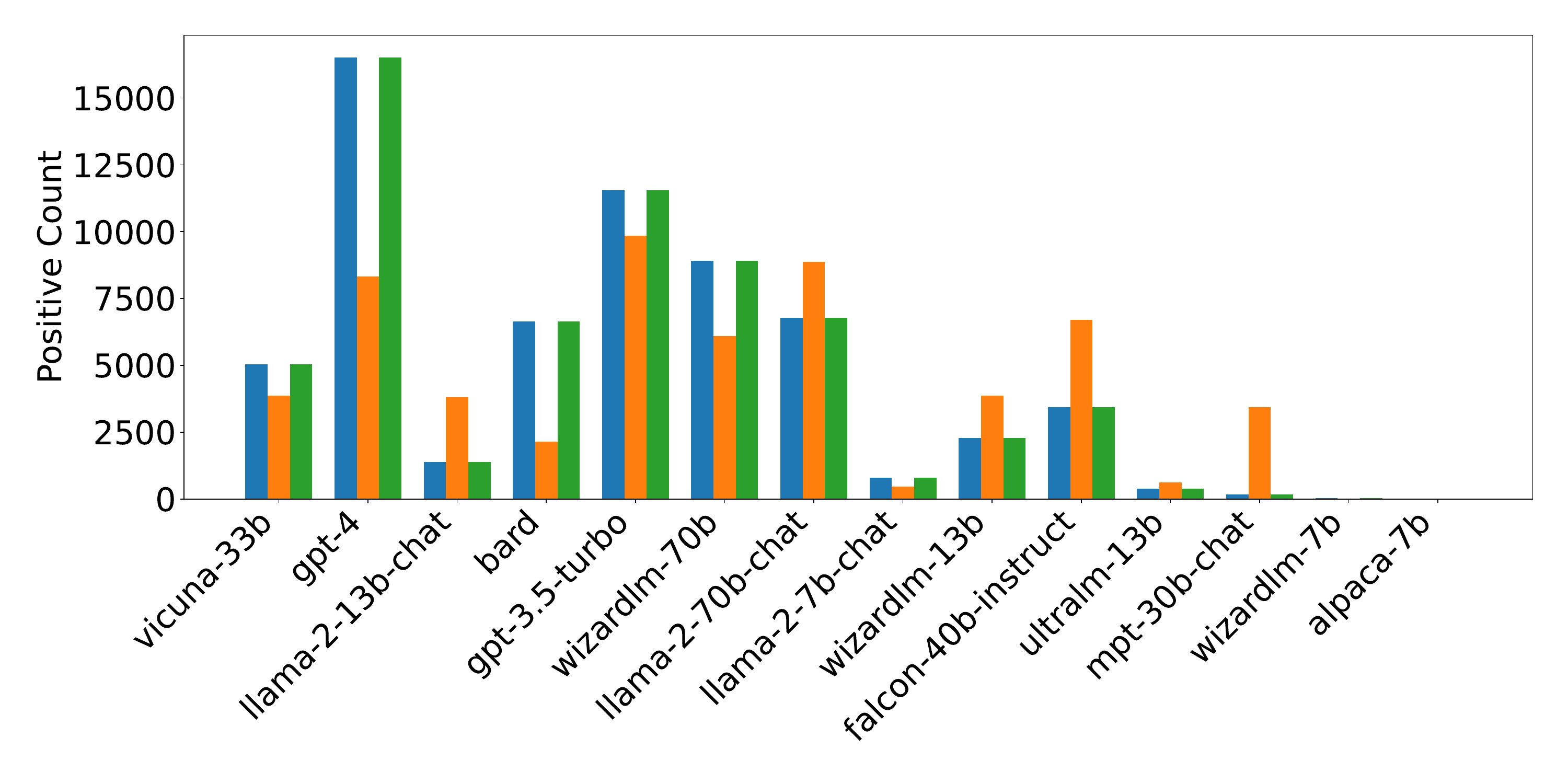}
    \caption{Positive}
\end{subfigure}
\begin{subfigure}{0.45\linewidth}
    \includegraphics[width=\linewidth]{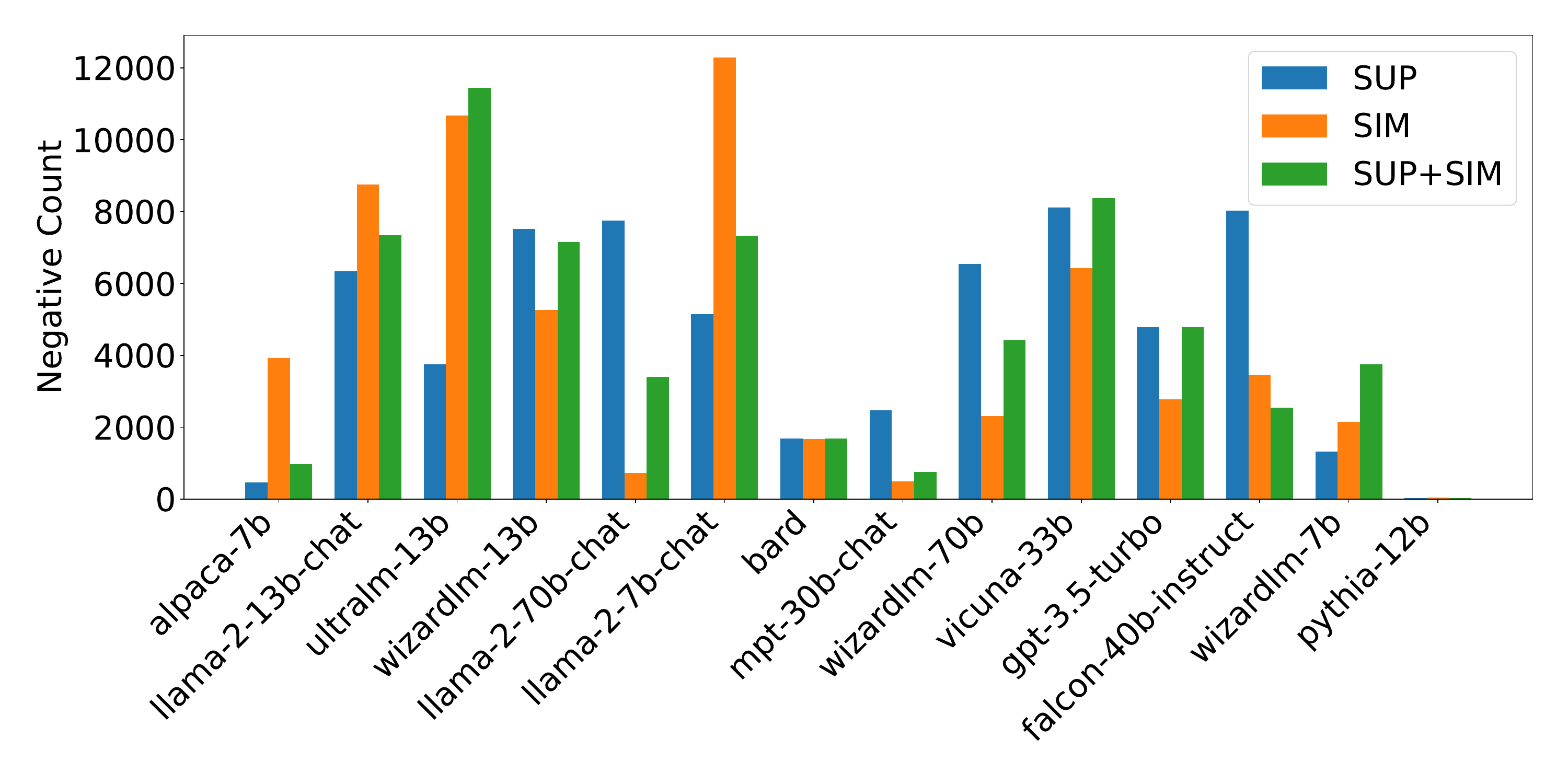}
    \caption{Negative}
\end{subfigure}
    \caption{Selected Model Frequency for the positive and negative pairs.}
    \label{fig:model_frequencies}
\end{figure*}

\begin{figure*}
    \centering
\begin{subfigure}{0.23\linewidth}
    \includegraphics[width=\linewidth, height=4cm]{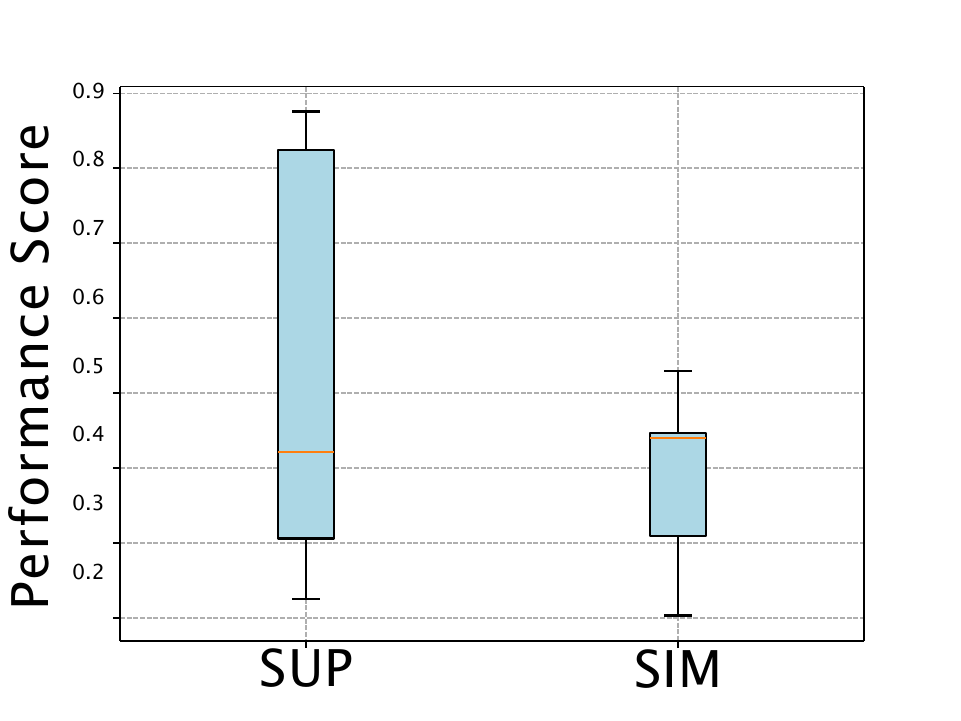}
    \caption{Qwen-2.5-3B}
\end{subfigure}
\begin{subfigure}{0.23\linewidth}
    \includegraphics[width=\linewidth, height=4cm]{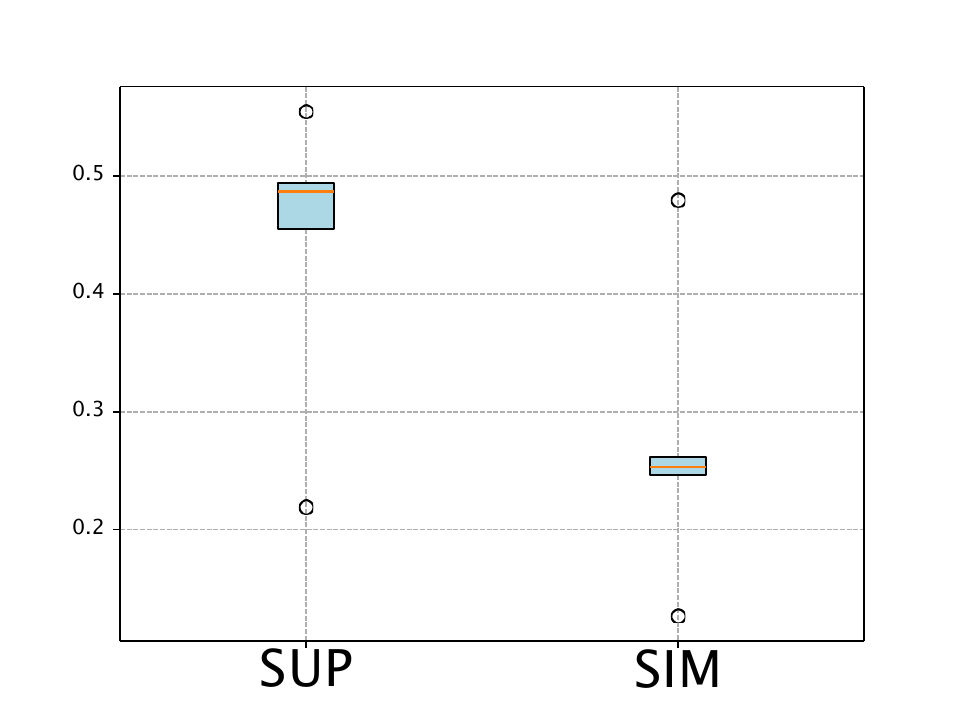}
    \caption{Llama-3.1-3B}
\end{subfigure}
\begin{subfigure}{0.23\linewidth}
    \includegraphics[width=\linewidth, height=4cm]{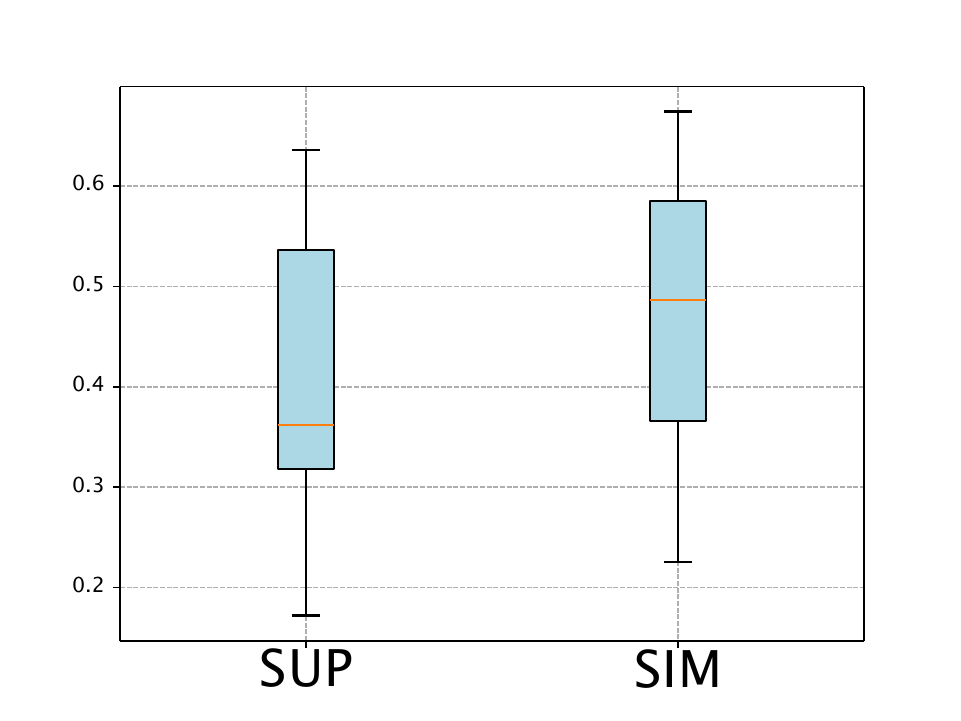}
    \caption{Qwen-2.5-7B}
\end{subfigure}
\begin{subfigure}{0.23\linewidth}
    \includegraphics[width=\linewidth, height=4cm]{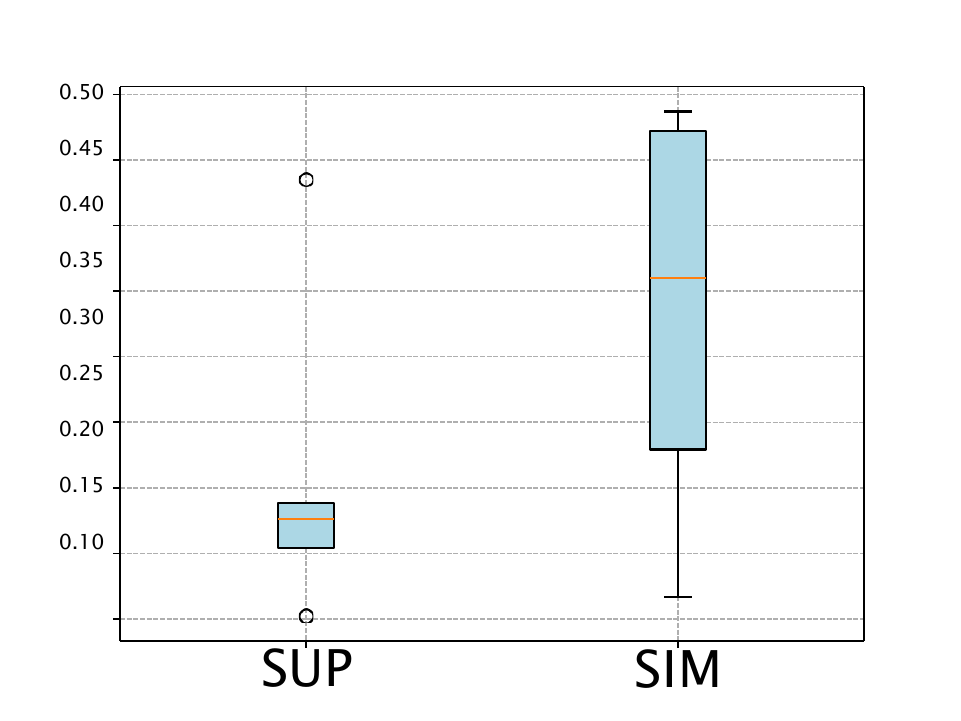}
    \caption{Llama-3.1-8B}
\end{subfigure}
    \caption{Visualization of performance across different models. This represents the performance when using the DPO training method. For smaller models, using SUP results in better performance, whereas for larger models, SIM yields better performance}
    \label{fig:model_size_kinship}
\end{figure*}

\begin{table*}[t]
\begin{adjustbox}{width=\textwidth}
\begin{tabular}{c|cccccc|c}
\toprule
     Model &  IFeval &  MMLU-STEM &  MMLU-pro &  Hellaswag &  ARC-easy &  ARC-Challenge &  Average Score \\
\midrule
     GPT-4 &        85.37 & 86.40 &      0.64 &      95.30 &     96.63 & 96.40 &  0.99 \\
      GPT-3.5-turbo &        72.54 & 70.00 &      0.38 &      85.00 &     92.80 & 83.02 &  0.77 \\
      bard &        89.33 & 71.80 &      0.59 &      84.70 &     84.43 & 77.13 &  0.86 \\
    Llama-2-7b-chat &        25.19 & 53.10 &      0.20 &      67.50 &     72.14 & 54.61 &  0.41 \\
   Llama-2-13b-chat &        24.82 & 57.80 &      0.25 &      71.20 &     72.05 & 58.02 &  0.45 \\
   Llama-2-70b-chat &        24.07 & 68.90 &      0.38 &      78.10 &     82.20 & 67.66 &  0.58 \\
        UltraLM-13b &        54.92 & 49.58 &      0.19 &      54.00 &     57.58 & 48.04 &  0.40 \\
        WizardLM-7b &        45.83 & 42.50 &      0.18 &      77.70 &     39.48 & 34.90 &  0.34 \\
       WizardLM-13b &        33.92 & 52.30 &      0.17 &      81.00 &     72.94 & 55.38 &  0.45 \\
       WizardLM-70b &        49.51 & 52.70 &      0.27 &      83.30 &     80.68 & 71.93 &  0.58 \\
         Vicuna-33b &        52.76 & 64.00 &      0.23 &      75.00 &     81.57 & 64.51 &  0.57 \\
 Alpaca-7b &        30.58 & 37.92 &      0.15 &      23.60 &     43.31 & 33.79 &  0.16 \\
Falcon-40b-instruct &        24.54 & 67.50 &      0.14 &      80.00 &     76.60 & 56.70 &  0.47 \\
       MPT-30b-chat &        30.70 & 50.40 &      0.20 &      24.53 &     87.12 & 70.73 &  0.38 \\
  Starchat &        28.30 & 40.12 &      0.12 &      25.40 &     16.96 &  9.07 &  0.05 \\
         Pythia-12b &        24.71 & 27.00 &      0.12 &      25.60 &     24.49 & 31.80 &  0.07 \\
\bottomrule
\end{tabular}
\end{adjustbox}
\caption{Benchmark Scores for Trained Models. Multiple-choice benchmarks (MMLU-STEM, HellaSwag, ARC-Easy, and ARC-Challenge) are evaluated based on accuracy. IFeval and MMLU-Pro are assessed using its own metric. The average score is computed after min-max normalization.}
\label{tab:benchmark_performance}
\end{table*}

\begin{figure*}
    \includegraphics[width=1\linewidth]{latex/figures/example_similarity.pdf}
    \caption{The process of calculating \( MB-SUP \) and \( MB_SIM \). Benchmark performance is normalized using min-max normalization, and the overall SIM is performed using cosine similarity.}
    \label{fig:example_sim}
\end{figure*}

\section{Benchmark Performance Representation}
What benchmark represent the model capability: knowledge, instruction-following, and reasoning?
Models exhibit distinct similarity patterns based on their capabilities, with these patterns varying across different evaluation metrics. Notably, model behavior knowledge is influenced by factors such as model scale and training methodology, leading to variations in clustering behavior across different capability dimensions.

\paragraph{Analysis Framework}
To systematically investigate these relationships, we analyzed model similarities across three key capability dimensions:

\begin{itemize}
    \item \textbf{Knowledge-Based Tasks}: Unlike reasoning capability, clustering in instruction-following tasks is more strongly aligned with \textit{model families} rather than size. This indicates that training methodology and architectural choices exert a greater influence on instruction adherence.
    
    \item \textbf{Reasoning Capability}: Models tend to cluster primarily based on parameter count, suggesting that \textit{model size} plays a dominant role in shaping reasoning performance.
    
    \item \textbf{Instruction-Following (IF) Tasks}:  A hierarchical influence pattern emerges, where:
    \begin{itemize}
        \item At the initial hierarchy, the \textit{model family} is the primary determinant.
        \item At the higher hierarchy, the \textit{model size} becomes a stronger predictor of performance.
    \end{itemize}
\end{itemize}

\begin{figure}[]
\centering
\begin{subfigure}{0.48\textwidth}
    \includegraphics[width=\linewidth]{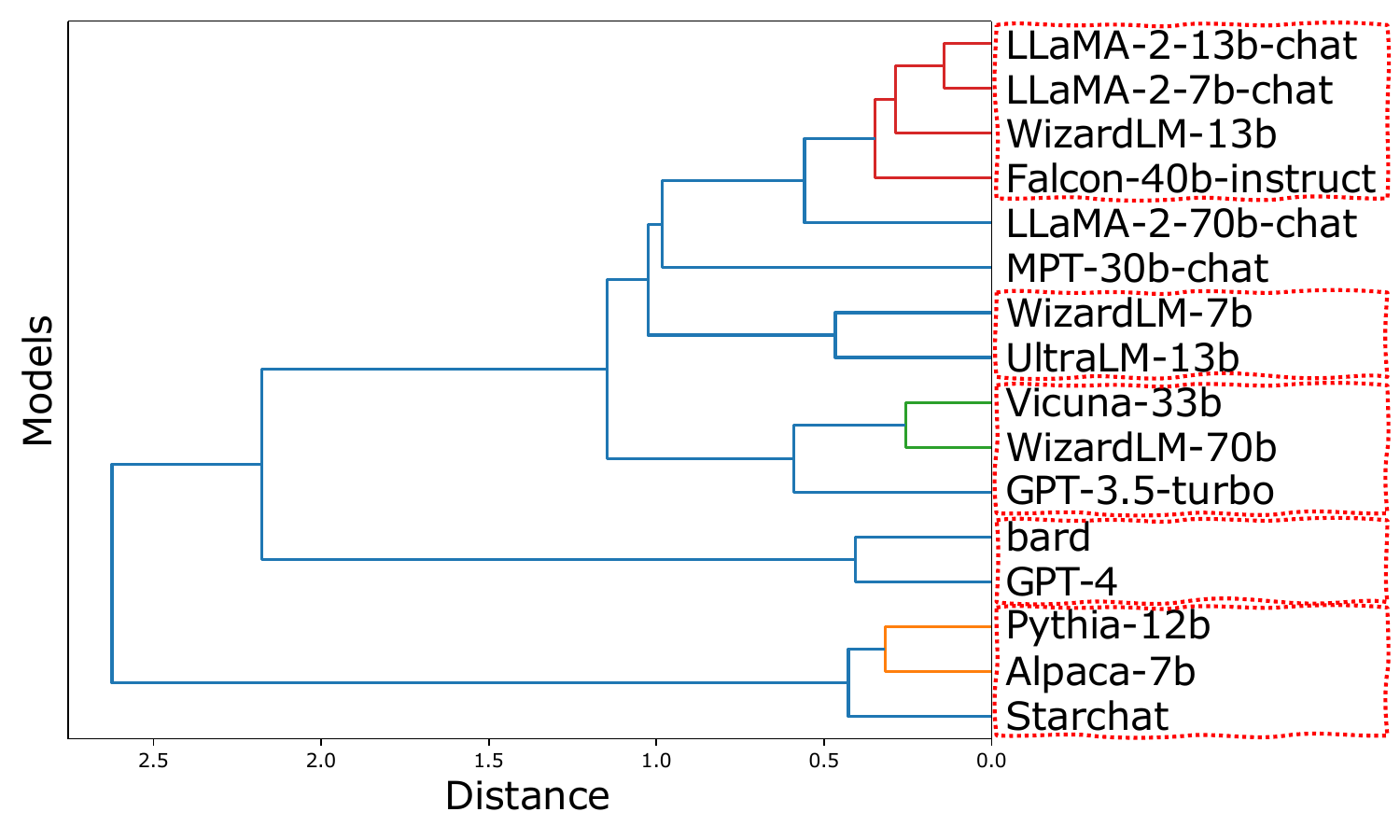}
    \caption{Total}
    \label{fig:Knowledge}
\end{subfigure}

\begin{subfigure}{0.48\textwidth}
    \includegraphics[width=\linewidth]{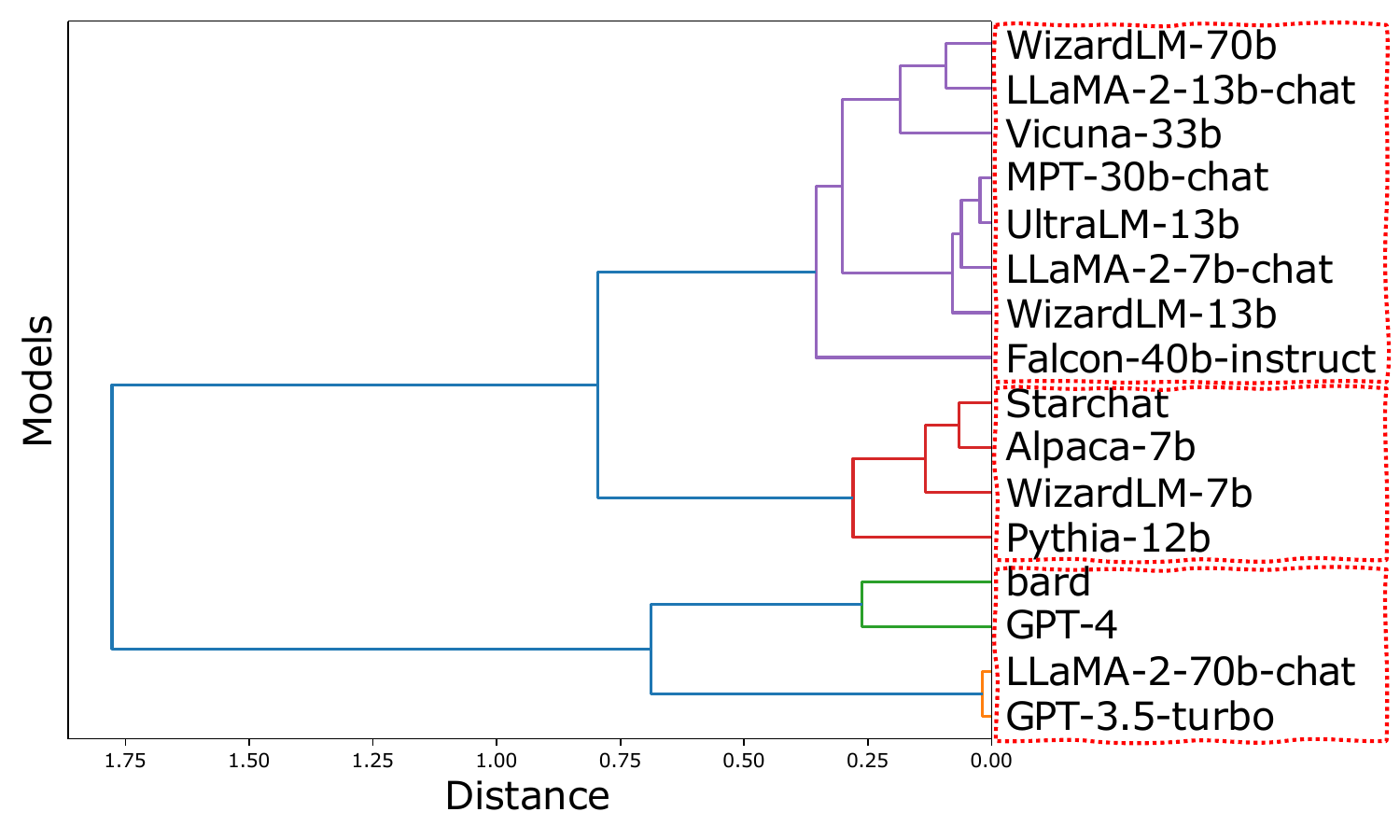}
    \caption{Knowledge}
    \label{fig:Knowledge}
\end{subfigure}

\begin{subfigure}{0.48\textwidth}
    \includegraphics[width=\linewidth]{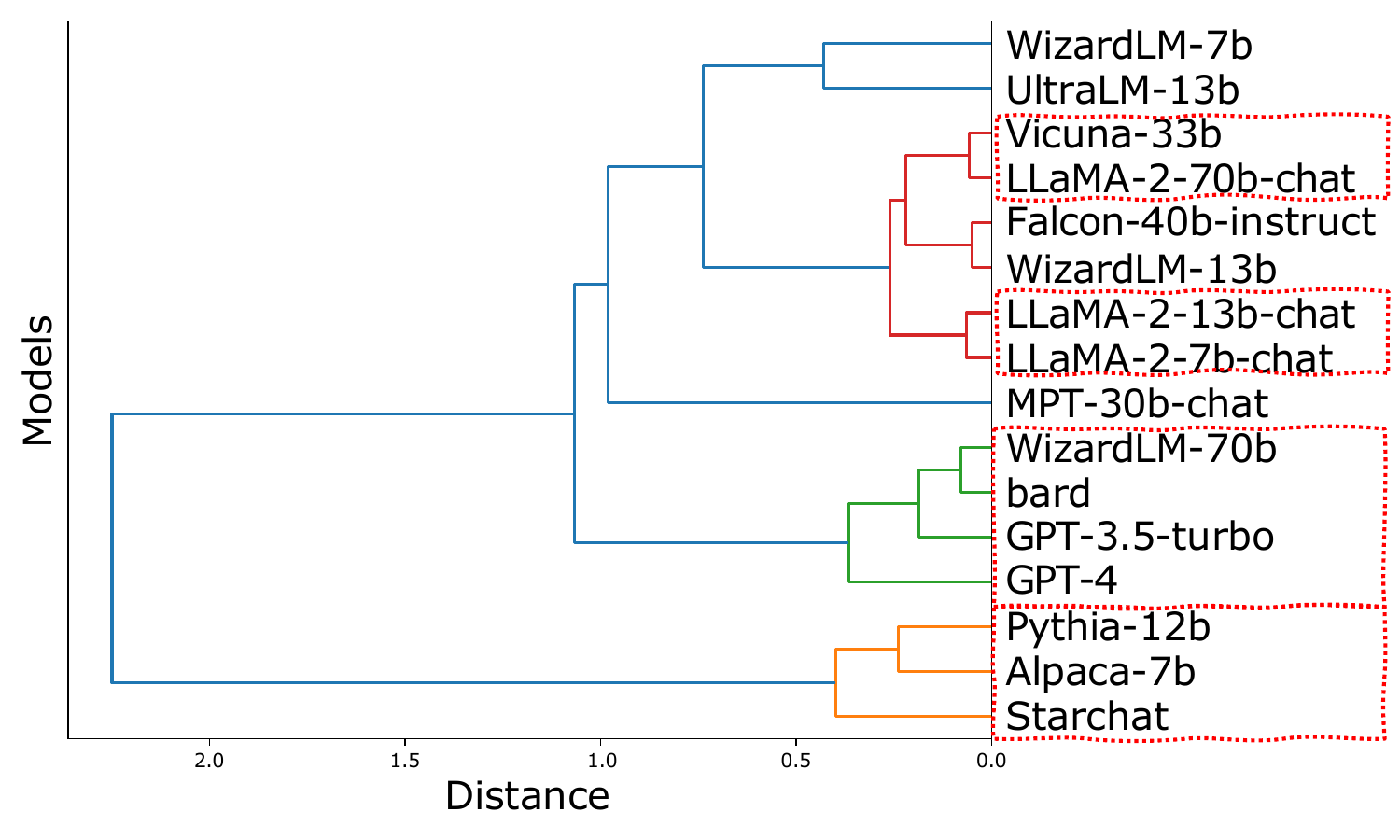}
    \caption{Reasoning}
    \label{fig:Reasoning}
\end{subfigure}

\begin{subfigure}{0.48\textwidth}
    \includegraphics[width=\linewidth]{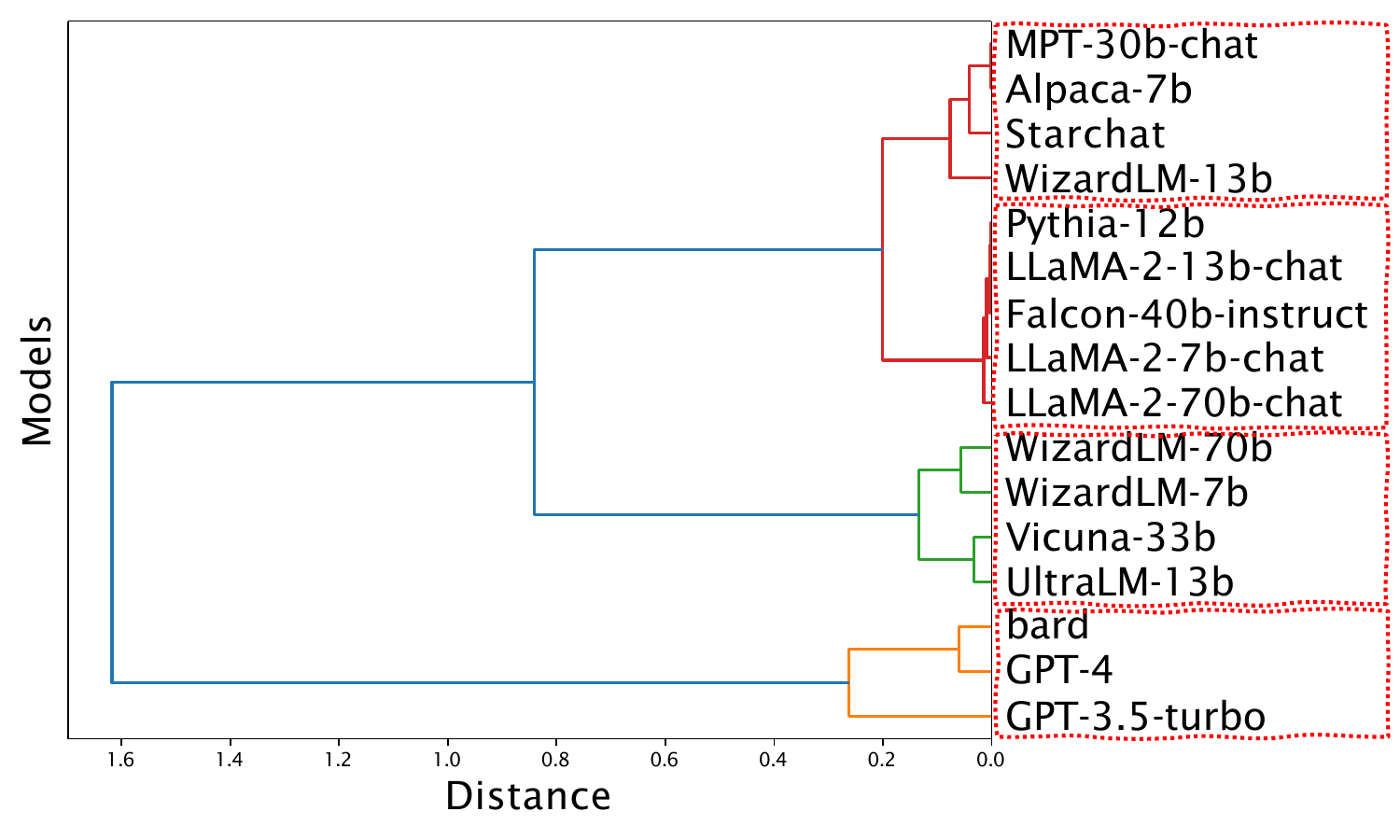}
    \caption{Instruction-following}
    \label{fig:Instruction_following}
\end{subfigure}
\caption{Dendrogram for Evaluation Categories: Knowledge, Reasoning, and Instruction-Following. Clusters exceeding a specific distance threshold (0.4) are highlighted in different colors. Models deemed similar share the same color line. The \textcolor{red}{red} dotted line shows the major groups of models in the dendrogram.}
\label{fig:benchmark_similarity}
\end{figure}

Figure~\ref{fig:benchmark_similarity} visualizes these relationships through dendrograms, illustrating the hierarchical clustering patterns that emerge across different capability dimensions.

Figure~\ref{fig:model_frequencies} shows the frequency of each model being selected as the positive or negative under each strategy.  
Overall, larger and more familiar models tend to be chosen more frequently.

\section{Total Performance}
\label{app:total_performance}
Table \ref{tab:total_performance} provides an aggregated view of the overall performance of each model across all benchmark datasets. The total performance scores were computed by averaging the normalized scores across the selected evaluation metrics, offering a holistic comparison of model capabilities. 



\section{Model Size and ZEBRA Strategies}
The impact of different Strategies varies significantly with model size, influencing how models learn from preference data. To investigate this relationship, we conducted experiments using models from the same family but with different parameter counts, specifically comparing small models (3B parameters) and larger models (7-8B parameters). Our analysis reveals a clear pattern in how model size determines the optimal strategy.

Smaller models (3B) exhibit superior performance when trained using the SUP algorithm, which prioritizes learning from response quality within the binarized dataset. In contrast, larger models (7-8B) achieve better results with the SIM algorithm, suggesting that response similarity becomes increasingly important as model size grows. 

This trend indicates that model size fundamentally influences how different models leverage preference data. While smaller models benefit more from explicit learning based on absolute quality differences, larger models demonstrate greater sensitivity to the nuanced relationships between similar responses in the training data. This relationship is visualized in Figure \ref{fig:model_size_kinship}, illustrating the distinct learning behaviors observed across different model scales.

\begin{table*}[]
    \centering
    \begin{tabular}{c|c|c|c|c|c|c}
\hline
Benchmark  & \multicolumn{1}{c|}{Model}    & \multicolumn{1}{c|}{} & Baseline & SUP        & SIM & SUP+SIM    \\ \hline
\multicolumn{1}{c|}{\multirow{8}{*}{MMLU-STEM}} & \multirow{2}{*}{Llama-3.1-3B} & SFT & 0.2331 & \textbf{0.2679} & 0.2434 & 0.2200 \\ \cline{3-7} 
\multicolumn{1}{c|}{}         &  & DPO & 0.3130 & \textbf{0.4868} & 0.2528 & 0.3940 \\ \cline{2-7} 
\multicolumn{1}{c|}{}         & \multirow{2}{*}{Llama-3.1-8B} & SFT & \textbf{0.2800} & 0.2460 & 0.2660 & 0.2240 \\ \cline{3-7} 
\multicolumn{1}{c|}{}         &  & DPO & \textbf{0.4902}   & 0.4349 & 0.4867 & 0.2880 \\ \cline{2-7} 
\multicolumn{1}{c|}{}         & \multirow{2}{*}{Qwen2.5-3B}   & SFT & \textbf{0.2760}   & 0.2200 & 0.2460 & 0.2500 \\ \cline{3-7} 
\multicolumn{1}{c|}{}         &  & DPO & \textbf{0.4706}   & 0.4212 & 0.4400 & 0.4580 \\ \cline{2-7} 
\multicolumn{1}{c|}{}         & \multirow{2}{*}{Qwen2.5-7B}   & SFT & \textbf{0.2800}   & 0.2280 & 0.2620 & 0.2780 \\ \cline{3-7} 
\multicolumn{1}{c|}{}         &  & DPO & \textbf{0.5120}   & 0.3620 & 0.4860 & 0.2800 \\ \hline

\multirow{8}{*}{MMLU-pro}      & \multirow{2}{*}{Llama-3.1-3B} & SFT & 0.1189   & 0.1230 & 0.1045 & \textbf{0.1332} \\ \cline{3-7} 
    &    & DPO & 0.1455   & \textbf{0.2193} & 0.1270 & 0.1516 \\ \cline{2-7} 
    & \multirow{2}{*}{Llama-3.1-8B} & SFT & 0.1004   & 0.1148 & 0.1045 & \textbf{0.1311} \\ \cline{3-7} 
    &    & DPO & \textbf{0.1168}   & 0.1025 & \textbf{0.1168} & 0.1107 \\ \cline{2-7} 
    & \multirow{2}{*}{Qwen2.5-3B}   & SFT & 0.1025   & 0.0840 & \textbf{0.1230} & 0.0820 \\ \cline{3-7} 
    &    & DPO & \textbf{0.2275}   & 0.2254 & 0.2029 & 0.2111 \\ \cline{2-7} 
    & \multirow{2}{*}{Qwen2.5-7B}   & SFT & 0.0861   & 0.1762 & \textbf{0.2275} & 0.1352 \\ \cline{3-7} 
    &    & DPO & \textbf{0.2377}   & 0.1721 & 0.2254 & 0.1700 \\ \hline
\multirow{8}{*}{IFeval}        & \multirow{2}{*}{Llama-3.1-3B} & SFT & \textbf{0.2494}   & 0.2410 & 0.2490 & 0.2206 \\ \cline{3-7} 
    &    & DPO & 0.3765   & \textbf{0.4940} & 0.4796 & 0.3033 \\ \cline{2-7} 
    
    & \multirow{2}{*}{Llama-3.1-8B} & SFT & 0.2494   & \textbf{0.4210} & 0.2470 & 0.2218 \\ \cline{3-7} 
    &    & DPO & 0.2421   & 0.1882 & 0.2292 & \textbf{0.2494} \\ \cline{2-7} 
    & \multirow{2}{*}{Qwen2.5-3B}   & SFT & \textbf{0.2190}   & 0.2134 & 0.2122 & 0.2050 \\ \cline{3-7} 
    &    & DPO & \textbf{0.3633}   & 0.3058 & 0.3094 & 0.1715 \\ \cline{2-7} 
    & \multirow{2}{*}{Qwen2.5-7B}   & SFT & \textbf{0.2290}   & 0.2134 & 0.2122 & 0.2083 \\ \cline{3-7} 
    &    & DPO & 0.3321   & 0.3177 & \textbf{0.3657} & 0.2407 \\ \hline    
\multirow{8}{*}{ARC-easy}      & \multirow{2}{*}{Llama-3.1-3B} & SFT & \textbf{0.2660}   & 0.2460 & 0.2000 & 0.2340 \\ \cline{3-7} 
    &    & DPO & 0.6111   & 0.5547 & 0.2618 & \textbf{0.6679} \\ \cline{2-7} 
    & \multirow{2}{*}{Llama-3.1-8B} & SFT & 0.2180   & 0.2330 & \textbf{0.2400} & 0.2360 \\ \cline{3-7} 
    &    & DPO & 0.2176   & 0.1540 & \textbf{0.4720} & 0.0680 \\ \cline{2-7} 
    & \multirow{2}{*}{Qwen2.5-3B}   & SFT & 0.2410   & \textbf{0.2560} & 0.2480 & \textbf{0.2560} \\ \cline{3-7} 
    &    & DPO & 0.8497   & \textbf{0.8754} & 0.5295 & 0.8157 \\ \cline{2-7}
    & \multirow{2}{*}{Qwen2.5-7B}   & SFT & 0.2550   & 0.2260 & 0.2280 & \textbf{0.2900} \\ \cline{3-7} 
    &    & DPO & 0.5700   & 0.6359 & 0.6738 & \textbf{0.7134} \\ \hline
    
\multirow{8}{*}{ARC-challenge} & \multirow{2}{*}{Llama-3.1-3B} & SFT & \textbf{0.2556}   & 0.2492 & 0.2266 & 0.2019 \\ \cline{3-7} 
    &    & DPO & 0.5122   & 0.4551 & 0.2466 & \textbf{0.5712} \\ \cline{2-7} 
    & \multirow{2}{*}{Llama-3.1-8B} & SFT & 0.2320   & 0.2297 & \textbf{0.2761} & 0.2268 \\ \cline{3-7} 
    &    & DPO & \textbf{0.5463}   & 0.1763 & 0.3596 & 0.2946 \\ \cline{2-7} 
    & \multirow{2}{*}{Qwen2.5-3B}   & SFT & \textbf{0.2645}   & 0.2483 & 0.2227 & 0.2343 \\ \cline{3-7} 
    &    & DPO & 0.7398   & \textbf{0.8241} & 0.4470 & 0.7078 \\ \cline{2-7} 
    & \multirow{2}{*}{Qwen2.5-7B}   & SFT & 0.2343   & \textbf{0.2552} & 0.2483 & 0.2390 \\ \cline{3-7} 
    &    & DPO & 0.4432   & 0.5358 & \textbf{0.5847} & 0.3870 \\ \hline

\end{tabular}
    \caption{Model Performance Comparisons on Knowledge, Instruction-Following, and Reasoning-Related Tasks. The baseline is Instance-wise RLAIF \cite{ultrafeedback}. "Llama 3.1" refers to the Llama-3.1-Instruct series, and "Qwen-2.5" refers to the Qwen2.5-Instruct series. The training methods include Supervised Fine-Tuning (SFT) and Direct Preference Optimization (DPO). The \textbf{bold} text indicates the best performance for each model and training method.}
    \label{tab:total_performance}
\end{table*}

\end{document}